\documentclass[runningheads,orivec]{llncs}

\usepackage[T1]{fontenc}

\usepackage{graphicx, multirow}
\usepackage{amsmath,amssymb,amsfonts}
\usepackage{makecell}

\usepackage[
colorlinks=true,
linkcolor=blue,
citecolor=blue,
urlcolor=blue
]{hyperref}

\usepackage[
backend=biber,
style=chem-acs
]{biblatex}

\AtBeginBibliography{\footnotesize}

\begin{document}

\title{Data Efficient Complex Feature Fusion Network For Hyperspectral Image Classification}

\author{
Maitreya Shelare \and
Atharva Satam \and
Poonam Sonar \and
Sneha Burnase
}

\institute{
Department of Electronics and Telecommunication,\\
Rajiv Gandhi Institute of Technology, University of Mumbai, India\\
\email{maitreya.cse@gmail.com},  
\email{atharvajsatam17@gmail.com}, 
\email{poonam.sonar@mctrgit.ac.in}, 
\email{sneharburnase@gmail.com}
}

\authorrunning{M. Shelare et al.}
\titlerunning{Data Efficient Complex Feature Fusion Network for HSI Classification}

\maketitle              
\begin{abstract}
This work presents a data-efficient variant of the Attention-Based Dual-Branch Complex Feature Fusion Network (CFFN) for hyperspectral image classification. The proposed model, termed DE-CFFN, retains the original two-stream structure: the Real-Valued Neural Network (RVNN) processes standard hyperspectral patches, while the Complex-Valued Neural Network (CVNN) handles their Fourier-transformed counterparts. The main contribution of this work lies in the feature extraction process and architectural enhancement. Factor Analysis is used for dimensionality reduction, offering improved latent feature representation over Principal Component Analysis. Additionally, both the RVNN and CVNN streams are structurally modified by successively halving the number of filters in the 3D convolutional layers to reduce complexity. The outputs of both branches are concatenated and passed through a Squeeze and Excitation (SE) block to enhance joint feature representation. Evaluated on the Pavia University and Salinas datasets, DE‑CFFN achieves classification performance comparable to CFFN, while significantly reducing model size, memory consumption, and inference latency, making it suitable for real‑time hyperspectral imaging applications.

\end{abstract}

\section{Introduction} \label{introduction}

Hyperspectral imaging has revolutionized remote sensing by capturing detailed spatial structures along with high-resolution spectral information. Each pixel contains a rich spectral signature spanning numerous narrow, contiguous bands, enabling precise tasks such as land cover classification and environmental monitoring~\cite{Tao2021}. However, the resulting high data dimensionality introduces significant computational and modeling challenges, commonly referred to as the Hughes phenomenon~\cite{kanthi-2020}.

Conventional machine learning algorithms, while offering interpretability, often fall short in modeling the complex spatial and spectral correlations inherent in hyperspectral image data. Their performance further deteriorates in low-label scenarios due to overfitting~\cite{ahmad-no-date,li-2019}. In contrast, deep learning models address these limitations by automatically learning hierarchical feature representations from raw data, enabling improved generalization and robustness. 

Among these, Convolutional Neural Networks (CNNs) have demonstrated strong performance in HSI classification by effectively extracting both spatial and spectral features in an end-to-end manner~\cite{7514991}. CNN-based models are typically categorized as 1D, 2D, or 3D depending on the dimensionality of the convolutions used. 1D models analyze spectral vectors for each pixel independently, capturing spectral dependencies but ignoring spatial context~\cite{mou-2017}. In contrast, 2D models treat HSI as standard images, extracting spatial patterns while overlooking spectral continuity~\cite{fang-2019}. 3D models apply volumetric convolutions over spatial-spectral patches, enabling joint feature learning across both domains at the cost of increased computational complexity~\cite{feng-2019}.

To balance accuracy and computational efficiency, several hybrid architectures have been proposed. Yu et al.~\cite{yu-2020} combined 2D convolutions for spatial encoding with lightweight 3D modules for spectral refinement. Roy et al. proposed HybridSN~\cite{roy-2020}, which employs 3D convolutions in the initial layers followed by 2D convolutions, achieving improved performance with reduced complexity. Li et al.~\cite{rs16162908} introduced a complex-valued CNN that integrates both 2D and 3D convolutions to more effectively capture spectral-spatial relationships. Peker et al. \cite{Peker2021} developed a classification framework combining CNN-based feature extraction with the complex-valued wavelet neural network classifier.

Alkhatib et al.~\cite{alkhatib-2023} proposed an attention-based deep learning model, which we term the Complex Feature Fusion Network (CFFN). The model integrates real-valued and complex-valued 3D convolutional streams in parallel to effectively capture spatial and spectral features from hyperspectral images. Although CFFN achieves strong classification performance, we aim to make it more efficient to improve its suitability for real-time hyperspectral applications.

To this end, we propose DE-CFFN, a more data efficient variant designed to maintain high classification accuracy while substantially reducing computational demands. The main contributions of this work are as follows:

\begin{enumerate}
  \item We replace Principal Component Analysis with Factor Analysis for dimensionality reduction, enabling the extraction of more discriminative latent features and enhancing the overall quality of representation.
  \item We introduce a progressive reduction in the number of filters across the 3D convolutional layers in both the real-valued and complex-valued processing streams, substantially lowering the overall computational complexity.
  \item We demonstrate that DE-CFFN achieves up to 72\% reduction in model parameters, 75\% reduction in memory usage, and 14\% reduction in inference time, while preserving classification accuracy across most evaluation scenarios.
\end{enumerate}

\section{Proposed Method} \label{method}
\begin{figure}
\includegraphics[width=\textwidth]{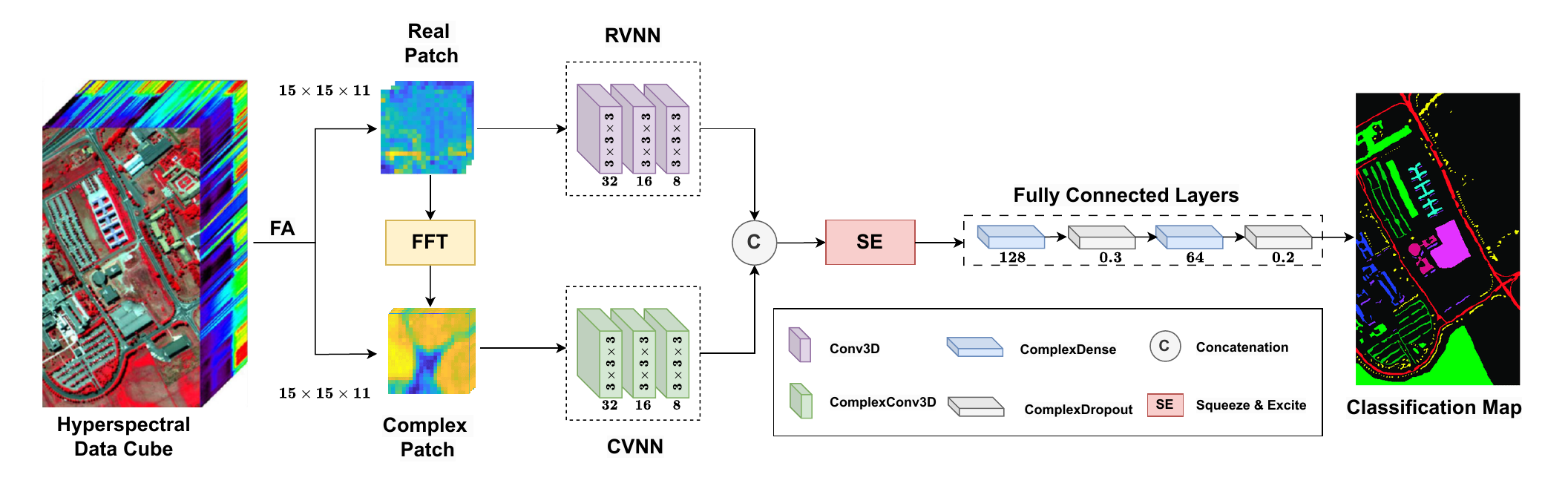}
\caption{Architecture of Data-Efficient Complex Feature Fusion Network (DE-CFFN)} \label{fig:architecture}
\end{figure}

The Data Efficient Complex Feature Fusion Network (DE-CFFN), shown in Fig.~\ref{fig:architecture}, is a dual-stream model designed for hyperspectral image classification. It jointly captures spatial, spectral, and frequency-domain features while maintaining computational efficiency.

To reduce spectral redundancy, Factor Analysis (FA) is applied to the input hyperspectral cube, producing a low-dimensional, discriminative representation. This is divided into overlapping patches of size $15 \times 15 \times d$, where $d$ is the number of retained spectral bands (empirically, $d = 11$). Each patch is simultaneously fed into a Real-Valued Neural Network (RVNN) and a Complex-Valued Neural Network (CVNN), forming the model’s two parallel streams.

The RVNN processes the real-valued patches using three successive 3D convolutional layers with decreasing filter sizes (32, 16, 8), each with kernel size $3 \times 3 \times 3$, stride 1, and ReLU activation. In parallel, the CVNN operates on the frequency-domain representation of the input, obtained via band-wise Fast Fourier Transform (FFT). It follows the same 3D convolutional structure but uses complex-valued convolutions and activation functions tailored to handle both magnitude and phase information.

After feature extraction, the RVNN output is cast to the complex domain and concatenated with the CVNN output along the channel dimension. This fused complex-valued feature map is then passed through seven stacked complex-valued Squeeze-and-Excitation (SE) blocks, each with a channel reduction ratio of 8. These attention modules adaptively recalibrate channel-wise dependencies, enhancing informative features while suppressing less relevant ones.

The refined tensor is flattened in the complex domain and passed through two fully connected complex-valued layers with 128 and 64 units, respectively, followed by dropout layers (0.3 and 0.1). Finally, classification is performed using a complex-valued dense layer with a real-valued softmax activation applied on the magnitudes of the outputs, yielding a pixel-wise classification map.

\subsection{RVNN Stream} \label{RVNN}
The RVNN stream operates on FA-reduced real-valued patches and leverages 3D convolutions to jointly capture spatial and spectral correlations, unlike 2D CNNs that treat spectral bands independently~\cite{Aburaed2021}. Its architecture consists of three layers with decreasing filters (32 $\rightarrow$ 16 $\rightarrow$ 8), ReLU activations, and kernel size $3 \times 3 \times 3$. This configuration balances expressive power and computational efficiency. The resulting features are cast to the complex domain for fusion with the CVNN stream.

\subsection{CVNN Stream} \label{CVNN}
The CVNN stream operates on frequency domain representations obtained via FFT along the spectral axis, producing complex-valued inputs that encode both amplitude and phase information. CVNNs effectively capture phase and amplitude relationships, which are crucial for hyperspectral image analysis~\cite{acar-2018, hirose-2012}.

This stream follows a structure similar to the RVNN, but utilizes complex convolutions and complex-specific activation functions such as Cartesian ReLU. This allows it to extract hierarchical features across the spectral, spatial, and frequency domains. When combined with the RVNN stream, the model benefits from complementary strengths, drawing spatial detail from the RVNN and frequency sensitivity from the CVNN~\cite{alkhatib-2023}.

\subsection{Squeeze-and-Excitation Attention Mechanism} \label{SE}
\begin{figure}[h]
\centering
\includegraphics[width=0.7\textwidth]{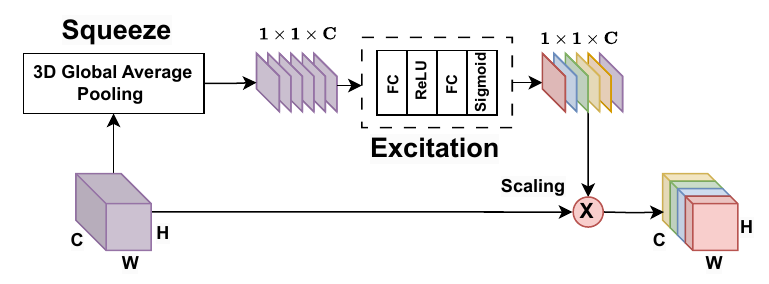}
\caption{Squeeze-and-Excitation (SE) Block.} \label{fig:SE}
\end{figure}

The Squeeze-and-Excitation (SE) mechanism~\cite{hu-2020} is a lightweight attention module that explicitly models inter-channel dependencies to recalibrate feature responses adaptively. It enhances the representational capacity of the network by allowing it to focus on the most informative channels while suppressing less useful ones. Within the DE-CFFN model, the SE mechanism is extended to the complex domain and is applied to the fused complex-valued feature maps obtained after combining the outputs from the RVNN and CVNN branches.

Let $U \in \mathbb{R}^{H \times W \times C}$ denote the intermediate feature map, where $H$ and $W$ are the spatial dimensions and $C$ is the number of channels. The squeeze step aggregates global spatial information for each channel using global average pooling:
\begin{equation}
z_{c} = \frac{1}{H \times W} \sum_{i=1}^{H} \sum_{j=1}^{W} u_{c}(i,j),
\label{Squeeze}
\end{equation}
resulting in a compact channel descriptor $z \in \mathbb{R}^{C}$ that summarizes the spatial context.

Next, the excitation step captures non-linear channel-wise dependencies using a bottleneck of two fully connected layers interleaved with a ReLU activation:
\begin{equation}
s = \sigma(W_2 \cdot \mathrm{ReLU}(W_1 z)),
\label{Excitation}
\end{equation}
where $W_1 \in \mathbb{R}^{\tfrac{C}{r} \times C}$ and $W_2 \in \mathbb{R}^{C \times \tfrac{C}{r}}$ are learnable weight matrices and $r$ is the channel reduction ratio (typically set to 8). The sigmoid activation $\sigma$ ensures that the resulting excitation vector $s \in \mathbb{R}^{C}$ scales each channel adaptively in the range [0, 1].

Finally, the recalibrated output $\widetilde{u}c$ is obtained by reweighting each input channel $u_c$ with its corresponding excitation value $s_c$:
\begin{equation}
\widetilde{u}{c} = s_c \cdot u_c.
\end{equation}

This channel-wise modulation allows the network to emphasize relevant features while downplaying noisy or redundant information.

\section{Results} \label{results}
We evaluate the DE-CFFN model on two widely used benchmark datasets, Pavia University and Salinas, to ensure a standardized assessment of its performance.

\begin{figure}[ht]
\centering
\includegraphics[width=0.7\textwidth]{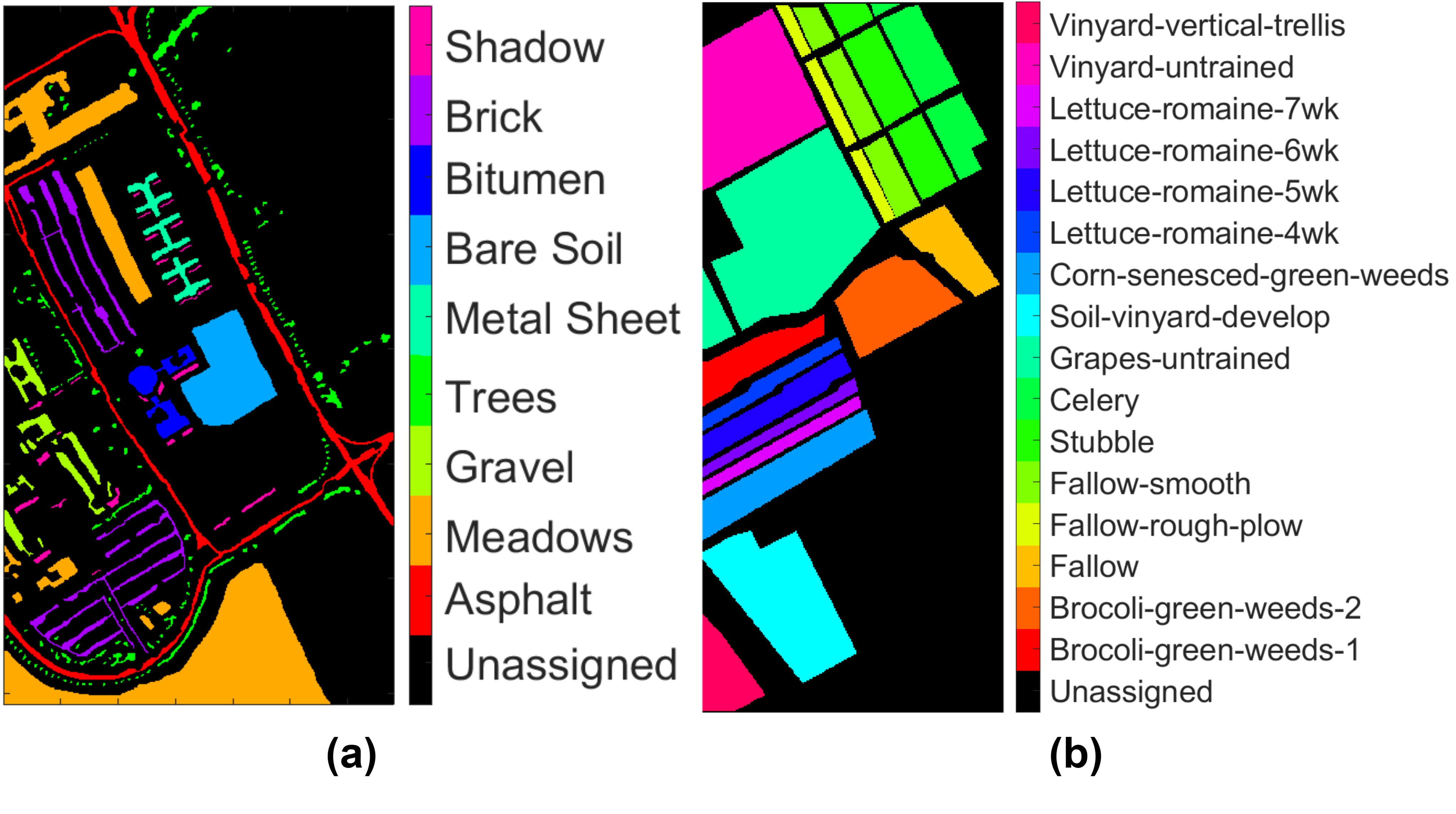}
\caption{Hyperspectral datasets: (a) Pavia University scene; (b) Salinas scene.} \label{fig:datasets}
\end{figure}

\subsection{Dataset Description} \label{Dataset Description}

The Pavia University (PU) dataset was acquired over an urban area in northern Italy using the ROSIS sensor. It offers a spatial resolution of 1.3 meters per pixel and includes ground truth annotations for nine land cover classes, predominantly covering urban structures and vegetation. The image spans approximately 793 × 441 meters.

The Salinas (SA) dataset was collected by the AVIRIS sensor over California’s Salinas Valley, a region known for agricultural diversity. With a spatial resolution of 3.7 meters per pixel, the scene includes sixteen annotated classes representing various crops and soil types. After removing 20 water absorption bands, 204 spectral bands were retained for analysis.

A summary of the datasets is provided in Table~\ref{tab:dataset}, and Figure~\ref{fig:datasets} displays RGB composites of both scenes along with their respective ground truth maps.

\begin{table}[ht]
\centering
\caption{Summary of Datasets}
\label{tab:dataset}
\begin{tabular}{|l|c|c|c|c|}
\hline
Dataset & Spatial Dimensions & Wavelength Range      & Spectral Bands & Classes \\
\hline
Pavia University (PU) & 610 × 340 pixels    & 430–860\,nm           & 103           & 9  \\
Salinas (SA)          & 512 × 217 pixels    & 360–2500\,nm           & 224$^{\dagger}$& 16 \\
\hline
\multicolumn{5}{l}{$^{\dagger}$20 water absorption bands removed for analysis.}
\end{tabular}
\end{table}
\subsection{Implementation Details}\label{Implementation Details}

We conducted experiments on both datasets using a 1\% training split, with the remaining 99\% reserved for testing. Patches were randomly sampled from the hyperspectral cubes. Based on preliminary tuning, optimal patch sizes of $9 \times 9$ for PU and $15 \times 15$ for SA were selected to balance contextual information with computational efficiency.

To ensure a fair comparison, all training hyperparameters were aligned with the original CFFN implementation by Alkhatib et al.~\cite{alkhatib-2023}. Factor Analysis was used for dimensionality reduction, retaining 11 latent spectral components across both datasets. The same preprocessed input pipeline was used for all baseline models to eliminate variations due to data handling.

The model was trained using the Adam optimizer with a learning rate of 0.001 and categorical cross-entropy loss. Training proceeded for up to 100 epochs with a batch size of 16, and early stopping based on validation loss was applied to mitigate overfitting. All models were evaluated using the same train–test splits across multiple runs to ensure statistical reliability.

All experiments were implemented in Python using the Keras API with a TensorFlow backend. Training and evaluation were conducted on a system equipped with an Intel Core i7-11800H CPU, 16 GB DDR4 RAM, and an NVIDIA GeForce RTX 3050 Ti GPU.

\subsection{Experimental Results}\label{Experimental Results}
The proposed DE-CFFN model was evaluated on the Pavia University and Salinas datasets against established HSI classification models, including SpectralNET~\cite{chakraborty2021spectralnet}, HybridSN~\cite{roy-2020}, and CFFN~\cite{alkhatib-2023}. To assess the impact of the Squeeze-and-Excitation (SE) block, both CFFN and DE-CFFN were tested with and without SE.

\begin{table}[t]
\centering

% --- Pavia University ---
\begin{minipage}{\textwidth}
\footnotesize
\centering
\caption{Classification accuracy of each class for the  Pavia University dataset obtained by SpectralNET~\cite{chakraborty2021spectralnet}, HybridSN~\cite{roy-2020}, CFFN~\cite{alkhatib-2023}, and proposed DE-CFFN.}
\label{PU_Table}
\resizebox{9cm}{!}{%
\begin{tabular}{|l|c|c|cc|cc|}
\hline
\multirow{2}{*}{Class} & \multirow{2}{*}{SpectralNET} & \multirow{2}{*}{HybridSN} & \multicolumn{2}{c|}{CFFN} & \multicolumn{2}{c|}{\textbf{DE-CFFN}} \\
\cline{4-7}
 & & & No SE & With SE & No SE & With SE \\
\hline
Asphalt      & 94.58 & \textbf{99.53} & 98.43 & 98.89 & 98.16 & 98.95 \\
Meadows      & 94.23 & \textbf{99.65} & 95.94 & 97.29 & 98.39 & 97.93 \\
Gravel       & 95.05 & 89.46 & 98.22 & \textbf{98.30} & 97.95 & 98.43 \\
Trees        & 94.71 & 89.12 & 98.86 & \textbf{99.28} & 98.81 & 98.78 \\
Metal Sheet  & 95.00 & 99.00 & \textbf{99.98} & \textbf{99.99} & \textbf{99.98} & \textbf{99.99} \\
Bare Soil    & 94.01 & 98.21 & 96.77 & 97.76 & 98.51 & \textbf{98.77} \\
Bitumen      & 94.87 & \textbf{99.63} & 99.46 & 99.68 & 99.30 & 99.51 \\
Brick        & 94.55 & 98.00 & 97.36 & 97.88 & 97.62 & \textbf{98.13} \\
Shadow       & 94.23 & 94.02 & \textbf{99.83} & 99.67 & 99.53 & 99.90 \\
\hline
OA (\%)      & 94.15 & 95.62 & 92.39 & 94.35 & 94.11 & \textbf{95.17} \\
AA (\%)      & \textbf{94.80} & 93.28 & 90.14 & 91.01 & 91.07 & 92.32 \\
Kappa$\times 100$ & 93.94 & 94.17 & 89.74 & 92.43 & 92.15 & \textbf{93.54} \\
\hline
\end{tabular}}
\end{minipage}

\vspace{1em} 

% --- Salinas ---
\begin{minipage}{\textwidth}
\footnotesize
\centering
\caption{Classification accuracy of each class for the Salinas dataset obtained by SpectralNET~\cite{chakraborty2021spectralnet}, HybridSN~\cite{roy-2020}, CFFN~\cite{alkhatib-2023}, and the proposed DE-CFFN.}
\label{SA_Table}
\resizebox{9.2cm}{!}{%
\begin{tabular}{|l|c|c|cc|cc|}
\hline
\multirow{2}{*}{Class} & \multirow{2}{*}{SpectralNET} & \multirow{2}{*}{HybridSN} & \multicolumn{2}{c|}{CFFN} & \multicolumn{2}{c|}{\textbf{DE-CFFN}} \\
\cline{4-7}
 & & & No SE & With SE & No SE & With SE \\
\hline
Brocoli-green-weeds-1     & 94.40 & 99.00 & 99.99 & \textbf{100.00} & 99.93 & 99.96 \\
Brocoli-green-weeds-2     & 95.00 & 98.86 & \textbf{100.00} & \textbf{100.00} & 99.98 & \textbf{100.00} \\
Fallow                    & 94.90 & 99.29 & 99.83 & \textbf{99.92} & \textbf{99.93} & \textbf{99.92} \\
Fallow-rough-plow         & 93.99 & \textbf{100.00} & 99.85 & \textbf{99.98} & 99.93 & 99.92 \\
Fallow-smooth             & 94.83 & 99.02 & 99.79 & \textbf{99.93} & 99.89 & 99.89 \\
Stubble                   & 95.00 & 99.00 & \textbf{100.00} & \textbf{100.00} & 99.99 & 99.99 \\
Celery                    & 94.60 & 99.92 & 99.99 & \textbf{100.00} & 99.97 & 99.92 \\
Grapes-untrained          & 94.48 & 99.04 & 97.23 & \textbf{97.68} & 96.93 & 96.83 \\
Soil-vinyard-develop      & 95.00 & 98.00 & \textbf{99.96} & 99.93 & 99.88 & 99.98 \\
Corn-senesced-green-weeds & 94.50 & \textbf{100.00} & 99.67 & 99.78 & 99.80 & 99.87 \\
Lettuce-romaine-4wk       & 94.39 & 99.50 & 99.83 & \textbf{99.96} & 99.84 & 99.76 \\
Lettuce-romaine-5wk       & 94.48 & 99.00 & 99.88 & \textbf{99.95} & 99.90 & 99.90 \\
Lettuce-romaine-6wk       & 94.57 & 98.81 & 99.78 & 99.86 & 99.84 & \textbf{99.95} \\
Lettuce-romaine-7wk       & 94.34 & 97.02 & 99.85 & 99.86 & 99.87 & \textbf{99.98} \\
Vinyard-untrained         & 93.85 & 98.83 & 97.24 & \textbf{97.66} & 96.97 & 96.87 \\
Vinyard-vertical-trellis  & 95.02 & 99.10 & 99.99 & \textbf{100.00} & 99.96 & 99.93 \\
\hline
OA (\%)                   & 94.83 & 96.05 & 96.42 & \textbf{97.25} & 96.31 & 96.34 \\
AA (\%)                   & 96.80 & 97.05 & 97.15 & \textbf{98.25} & 97.48 & 97.67 \\
Kappa$\times 100$        & 94.24 & 95.60 & 96.01 & \textbf{96.94} & 95.89 & 95.93 \\
\hline
\end{tabular}}
\end{minipage}

\end{table}

Evaluation metrics include Overall Accuracy (OA), Average Accuracy (AA), and the Kappa coefficient ($\kappa$). OA reflects the overall pixel-level classification accuracy, AA measures the average accuracy across all classes, and $\kappa$ accounts for agreement occurring by chance. Higher values across these metrics indicate better performance, with bold highlighting the best results.

As reported in Tables~\ref{PU_Table} and~\ref{SA_Table}, DE-CFFN consistently outperforms SpectralNET and HybridSN on both datasets.

On the Pavia University dataset, DE-CFFN with SE achieves the highest performance, with 95.17\% OA, 92.32\% AA, and a Kappa score of 93.54. For the Salinas dataset, it reaches 96.34\% OA, 97.67\% AA, and a Kappa score of 95.93, closely aligning with the top-performing CFFN variant.

The inclusion of SE generally improves classification accuracy, particularly for spectrally similar or minority classes. Notably, DE-CFFN without SE also demonstrates strong performance, underscoring the effectiveness of the dual-branch design.

\subsection{Complexity Analysis}\label{Complexity Analysis}

For real-world applications, hyperspectral classifiers must balance predictive performance with computational efficiency. To evaluate this trade-off, we compare DE-CFFN and CFFN in terms of model complexity (parameter count, memory usage, inference time) and classification metrics (OA, AA, and $\kappa$) across the PU and SA datasets.

\renewcommand{\arraystretch}{1.3}  
\setlength{\tabcolsep}{4pt}     

\begin{table}[ht]
\centering
\caption{Comparison of CFFN~\cite{alkhatib-2023} and DE-CFFN on PU and SA datasets. Bold indicates better OA, AA, $\kappa$, and lower complexity (parameters, memory, time).}
\label{tab:complexity}
\begin{tabular}{|c|c|c|c|c|c|c|c|}
\hline
Dataset & Model & \makecell{OA \\ (\%)} & \makecell{AA \\ (\%)} & \makecell{Kappa \\ ($\kappa\times100$)} & \makecell{Parameters \\ (M)} & \makecell{Memory \\ (MB)} & \makecell{Inference \\ Time (ms)} \\
\hline
\multirow{2}{*}{PU} 
        & CFFN      & 94.35   & 92.88   & 92.7    & 0.924          & 3.52        & \textbf{11.72}       \\
        & DE-CFFN   & \textbf{95.17}   & \textbf{92.32}   & \textbf{93.54}   & \textbf{0.257}   & \textbf{0.98} & 12.15         \\
\hline
\multirow{2}{*}{SA} 
        & CFFN      & \textbf{97.25}   & \textbf{96.71}   & \textbf{97.1}   & 6.823          & 26.03       & 35.06          \\
        & DE-CFFN   & 96.34   & 97.67   & 95.93    & \textbf{1.732}   & \textbf{6.61} & \textbf{30.27}         \\
\hline
\end{tabular}
\end{table}

As summarized in Table~\ref{tab:complexity}, on the PU dataset, DE-CFFN achieves an OA of 95.17\% and $\kappa$ of 93.54, outperforming CFFN by 0.82\% and 0.84 points, respectively. Despite this improvement, DE-CFFN reduces the number of parameters by 72.2\% (from 0.924M to 0.257M) and memory usage by 72.2\% (from 3.52 MB to 0.98 MB). The inference time remains nearly identical, increasing only slightly from 11.72 ms to 12.15 ms. Although average accuracy (AA) drops marginally by 0.56\%, the model still achieves competitive class-wise performance, suggesting effective generalization even under reduced complexity.

On the SA dataset, DE-CFFN achieves an AA of 97.67\%, which is 0.96\% higher than CFFN. While its OA and $\kappa$ decrease by 0.91\% and 1.17 points respectively, DE-CFFN offers considerable computational advantages, reducing parameters by 74.6\% (from 6.823M to 1.732M), memory usage by 74.6\% (from 26.03 MB to 6.61 MB), and inference time by 13.6\% (from 35.06 ms to 30.27 ms).

\subsection{Ablation Study}\label{Ablation Study}

To assess the impact of the dimensionality reduction technique on classification performance, an ablation study was performed on the DE-CFFN model with SE using Factor Analysis (FA) and Principal Component Analysis (PCA), each retaining 11 components for parity.

\renewcommand{\arraystretch}{1.3}  
\setlength{\tabcolsep}{4pt}     

\begin{table}[ht]
\centering
\caption{Ablation study on DE-CFFN with SE: Comparison of Factor Analysis (FA) and Principal Component Analysis (PCA) on PU and SA datasets. Bold indicates superior results.}
\label{tab:ablation}
\begin{tabular}{|c|c|c|c|c|c|c|}
\hline
\multirow{2}{*}{Method} 
& \multicolumn{3}{c|}{PU}      
& \multicolumn{3}{c|}{SA}      \\
\cline{2-7}
& \makecell{OA \\ (\%)} 
& \makecell{AA \\ (\%)} 
& \makecell{Kappa \\ ($\kappa\times100$)} 
& \makecell{OA \\ (\%)} 
& \makecell{AA \\ (\%)} 
& \makecell{Kappa \\ ($\kappa\times100$)} \\
\hline
PCA & 93.83 & 89.75 & 91.75 & 94.23 & 91.37 & 93.56 \\
FA  & \textbf{95.17} & \textbf{92.32} & \textbf{93.54} 
    & \textbf{96.34} & \textbf{97.67} & \textbf{95.93} \\
\hline
\end{tabular}
\end{table}

As shown in Table~\ref{tab:ablation}, FA consistently outperforms PCA on both the Pavia University (PU) and Salinas (SA) datasets. On PU, it improves OA by 1.34\%, AA by 2.57\%, and Kappa by 1.79. On SA, the gains are even more pronounced, with respective increases of 2.11\%, 6.30\%, and 2.37.

These results indicate that FA better captures latent spectral dependencies and reduces inter-band redundancy, enabling the model to learn more compact and discriminative features for hyperspectral image classification.

\section{Conclusion} \label{conclusion}

This paper introduced DE-CFFN, a data-efficient dual-branch model designed for real-time hyperspectral image (HSI) classification. The model combines a real-valued neural network (RVNN) for learning spatial-spectral features with a complex-valued neural network (CVNN) that processes frequency-domain information via Fourier transforms. To improve feature compactness, Factor Analysis (FA) is used for dimensionality reduction, while a Squeeze-and-Excitation (SE) block helps the network focus on more informative channels. Experiments on the Pavia University and Salinas datasets show that DE-CFFN delivers strong classification performance with up to 75\% fewer parameters and lower memory usage. This efficiency makes it ideal for real-time applications, such as onboard processing in UAVs or satellites, where fast and accurate inference matters more than resource constraints. In the future, we plan to scale the model to larger datasets, explore integration with temporal data, and test its usefulness across other remote sensing tasks.

\printbibliography
\end{document}